\documentclass[runningheads]{llncs}

\usepackage{graphicx}
\usepackage{comment}
\usepackage{amsmath,amssymb}
\usepackage{color}
\usepackage{url}
\usepackage{hyperref}
\usepackage{graphicx}
\usepackage{makecell}
\usepackage{cite}
\usepackage{wrapfig}
\usepackage[table]{xcolor}

\usepackage{caption} 
\usepackage{flafter} 

\usepackage{subcaption}

\newif\ifreview
\reviewtrue
\reviewfalse

\ifreview
	\usepackage{lineno}

	\linenumbers
\fi

\begin{document}


\def\SubNumber{074}

\def\GCPRTrack{Main Track}

\title{On the Robustness of 3D Object Detectors}
\def\thefootnote{*}\footnotetext{These authors contributed equally to this work.}

\ifreview
\else

	\author{Fatima Albreiki\thefootnote{*}\orcidID{0000-0002-4212-1017} \and
	Sultan Abughazal\thefootnote{*}\orcidID{0000-0003-0416-5964} \and
	Jean Lahoud\orcidID{0000-0003-0315-6484} \and
	Rao Anwer\orcidID{0000-0002-9041-2214} \and
	Hisham Cholakkal\orcidID{0000-0002-8230-9065} \and
	Fahad Khan\orcidID{0000-0003-3449-7978}}

	\authorrunning{F. Albreiki et al.}
	
	\institute{Mohamed Bin Zayed University of Artificial Intelligence, Abu Dhabi, UAE\\
	\email{\{fatima.albreiki,sultan.abughazal,jean.lahoud,rao.anwer,\\hisham.cholakkal,fahad.khan\}@mbzuai.ac.ae}}
\fi

\maketitle              

\begin{abstract}
In recent years, significant progress has been achieved for 3D object detection on point clouds thanks to the advances in 3D data collection and deep learning techniques. Nevertheless, 3D scenes exhibit a lot of variations and are prone to sensor inaccuracies as well as information loss during pre-processing. Thus, it is crucial to design techniques that are robust against these variations. This requires a detailed analysis and understanding of the effect of such variations. This work aims to analyze and benchmark popular point-based 3D object detectors against several data corruptions. To the best of our knowledge, we are the first to investigate the robustness of point-based 3D object detectors. To this end, we design and evaluate corruptions that involve data addition, reduction, and alteration. We further study the robustness of different modules against local and global variations. Our experimental results reveal several intriguing findings. For instance, we show that methods that integrate Transformers at a patch or object level lead to increased robustness, compared to using Transformers at the point level. 





\end{abstract}
\section{Introduction}
\begin{figure}[t!]
     \centering
     \begin{subfigure}[b]{0.24\textwidth}
         \centering
         \includegraphics[width=\textwidth]{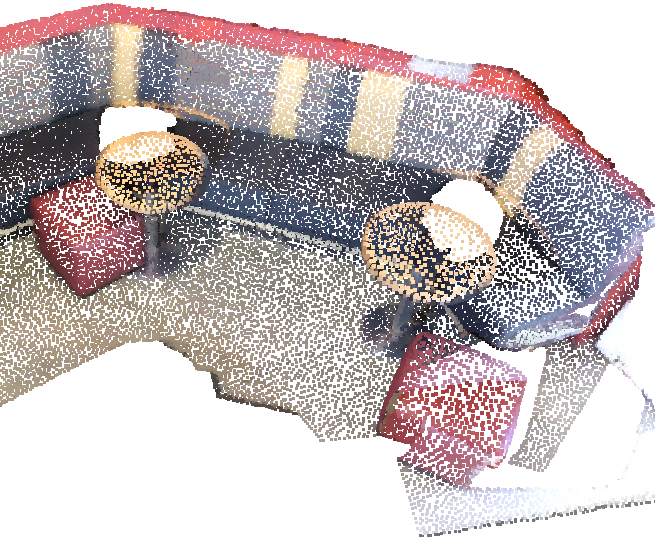}
         \caption{Original Scene}
     \end{subfigure}
     \hfill
     \begin{subfigure}[b]{0.24\textwidth}
         \centering
         \includegraphics[width=\textwidth]{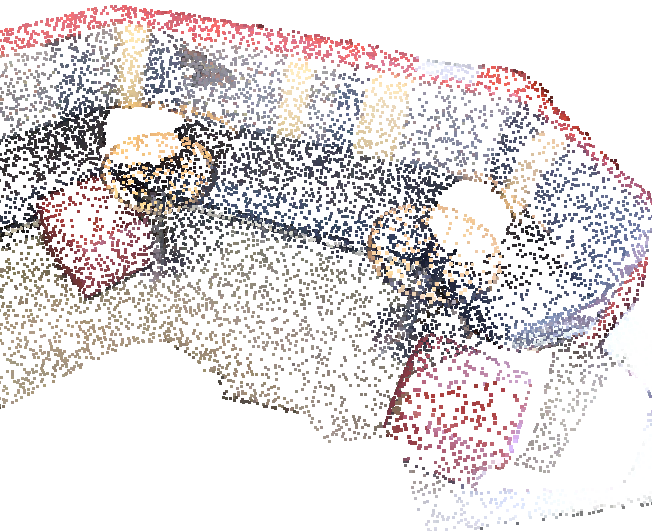}
         \caption{Point Removal}
     \end{subfigure}
     \hfill
     \begin{subfigure}[b]{0.24\textwidth}
         \centering
         \includegraphics[width=\textwidth]{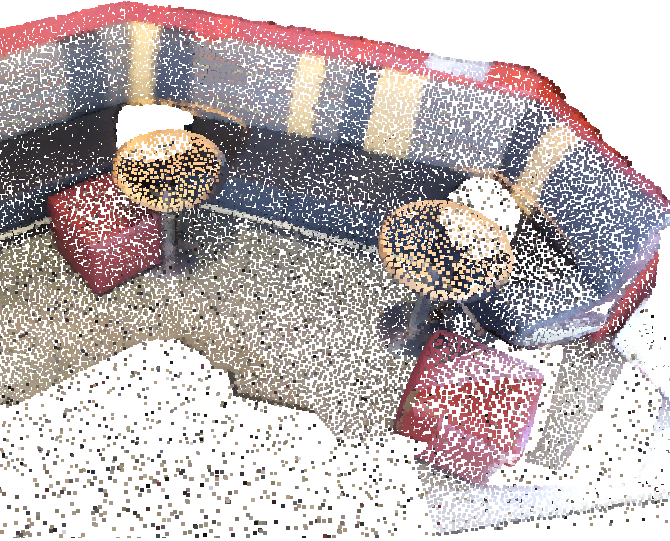}
         \caption{Point Addition}
     \end{subfigure}
     \hfill
     \begin{subfigure}[b]{0.24\textwidth}
         \centering
         \includegraphics[width=\textwidth]{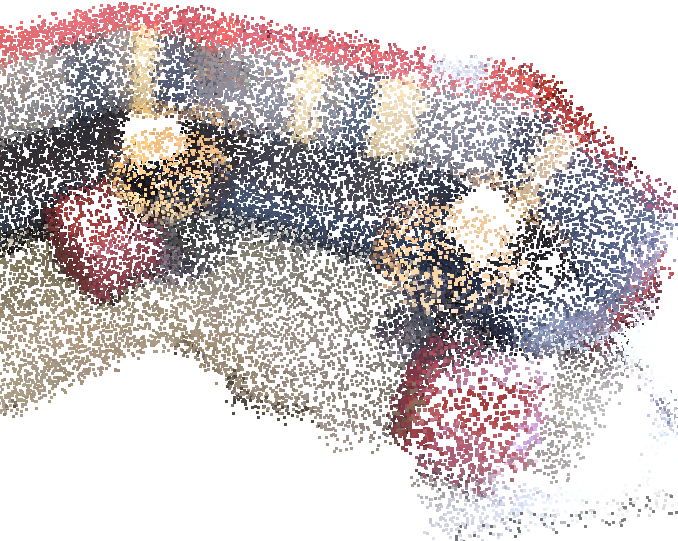}
         \caption{Point Alteration}
     \end{subfigure}\vspace{-0.1cm}
        \caption{We design point cloud corruptions and study their effect on 3D object detection. We group our corruptions into three categories: (a) corruptions that involve removing  points from the scene, (b) corruptions that add points to the scene, or (c) corruptions that alter the points within the scene.}\vspace{-0.3cm}
        \label{fig:overview}
\end{figure}

3D object detection aims at localizing objects of interest in a given 3D scan. It has applications in numerous areas, such as autonomous driving, robotics, and augmented reality. 3D object detection involves placing 3D bounding boxes around objects, which allows the understanding of real physical dimensions and distances to objects, compared to 2D image detection in which understanding is limited to the image plane. 

Most 3D object detection methods rely on the availability of depth information, which is usually acquired using range sensors, such as LiDAR for outdoors and Kinect for indoors. The accuracy of the acquired depth is greatly affected by sensor type and distance to the sensor. Moreover, camera viewpoint and scene structure lead to different representations of the same object, mainly due to occlusions. Also, scene acquisition is sometimes collected from different viewpoints, which requires aggregation of information using various reconstruction techniques, leading to more variations in how a given scene can be represented.



3D scenes are usually represented by point clouds, which are sets of unordered points. These point clouds are typically processed with deep learning methods that are either projection-based, voxel-based, or point-based. Projection-based methods project 3D data into a 2D plane and benefit from the advancement of image feature learning techniques. Voxel-based methods first transform the data into a regular grid to then be processed by 3D convolution neural networks (CNNs). Point-based methods directly process raw points, using MLP architectures \cite{pointnet,pointnetpp2017}, Graph convolutions \cite{wang2019dynamic}, and, recently, Transformers \cite{vaswani2017attention}.


 Point-based object detectors have shown promising performance on various datasets \cite{votenet2019,mlcvnet2020,groupfree3d2021,3detr2021}. Given the sparse nature of point clouds, point-based methods are generally less voluminous when compared to voxel-based and projection-based methods that include empty-space representation and suffer from quantization artifacts. Nonetheless, the robustness of point-based 3D object detection methods against point cloud variations is yet to be investigated. We, therefore, design corruptions to model variations in the point cloud capturing process. We also analyze and benchmark the performance of popular point-based 3D detection methods against the proposed corruptions.

Corruption in 3D points clouds can occur in various forms. We group corruptions into three main categories: (1) corruptions that lead to information loss, (2) corruptions from added information, and (3) corruptions due to data alteration. Examples of these corruptions are shown in Fig. \ref{fig:overview}. In our analysis, we simulate those sources of data corruption by generating multiple possible patterns in which each corruption can occur. We design seven point removal corruptions that represent information loss; namely, drop global, drop local, drop object, drop non-object, drop object parts, and drop floor. For the point addition corruptions, we introduce three different patterns: add global, add local, and scene expansion. Additionally, we simulate point alteration following four patterns: point jitter, background noise, local noise, and floor plane inclination. With our proposed corruptions, we strive to cover a broad variety of variations that are encountered in point cloud generation.


\textbf{Contributions:} We design point cloud corruptions to simulate variations due to errors in data collection and preparation. We analyze and benchmark the robustness of point-based 3D object detection architectures against these corruptions in point clouds. To the best of our knowledge, we are the first to investigate the robustness of point-based 3D object detectors. We also analyze the robustness of methods that use Transformers, which is the new paradigm in numerous computer vision applications including 3D object detection. Our experiments reveal several intriguing findings. For instance, we observe that the recently introduced transformer-based 3D detector shows increased robustness at the patch or object level, whereas its robustness diminishes when extracting point-level features.




\section{Related Work}

    
\noindent \textbf{3D Object Detection From Point Clouds.}    
    Many works have been proposed for the 3D object detection task. Most of these works build upon architectures that process 2D projections, voxels, or points. Projection-based methods reduce the computation cost by projecting 3D information into the 2D space \cite{yang2018pixor,wang2020pillar,chen2017multi}. Voxel-based methods transform the data into a regular grid, which can be processed with 3D convolution operations \cite{zhou2018voxelnet,yan2018second,shi2020pv}. Point-based methods directly process raw points to extract feature information. Many of these methods rely on PointNet \cite{pointnet} or PointNet++ \cite{pointnetpp2017}, which are MLP-based architectures that can directly operate on 3D points.
    VoteNet \cite{votenet2019} uses PointNet++ as a backbone to learn point features and Hough voting to identify object centers and generate object proposals. MLCVNet \cite{mlcvnet2020} is an extension of VoteNet, in which three context modules are added to exploit context information at the patch, object, and global level. Liu et al. \cite{groupfree3d2021} introduce a group-free point-based detection method that uses PointNet++ to extract features, and employs self-attention and cross-attention to extract and refine object features. 3DETR \cite{3detr2021} is an end-to-end transformer detection model that uses an encoder to encode and refine input features and a decoder to predict bounding boxes.

 \noindent \textbf{Robustness Benchmarks.}   
    Numerous robustness benchmarks have been presented for image classification. Hendrycks and Dietterich \cite{benchmarking2019} introduce ImageNet-C which consists of 15 corruptions applied to the validation set of ImageNet \cite{deng2009imagenet}. ImageNet-R \cite{benchmarkingHynd} enables the study of classifiers’ robustness to abstract visual renditions, attributes shift, and blurriness. Robustness benchmarks have been also proposed for 2D object detection methods. Michaelis \textit{et al.} \cite{ObjectDetectionBenchmark1} develop three corrupted versions of the popular object detection datasets, Pascal VOC \cite{everingham2010pascal}, MS COCO \cite{lin2014microsoft}, and Cityscapes \cite{cordts2016cityscapes}, to assess object detectors' robustness under different corruptions. Mirza \textit{et al.} \cite{ObjectDetectionBenchmark2} study the robustness of object detectors against degrading weather conditions. Robustness against corruptions was also studied in the 3D domain, mainly for the classification task. ModelNet40-C \cite{benchmarking2022_2} includes corruptions of ModelNet40 \cite{wu20153d} validation set with 15 common corruptions including Gaussian noise and occlusion. It provides comprehensive analysis of six model architectures performance with the proposed benchmark. Ren \textit{et al.} \cite{benchmarking2022} design and apply seven corruptions on ModelNet40 validation set. The benchmark includes 14 classification models and provides a systematic investigation of performance under corruptions.

\section{Point Cloud Corruptions}

\begin{table}
\setlength{\tabcolsep}{1mm} 
        \vspace{-1.5cm}
        \centering
         \caption{The proposed data corruptions and their causes. We design corruptions to simulate variations emerging from sensor accuracy (sens. acc.), sensor resolution (sens. res), sensor location (sens. loc.), scene variations (scene var.), and pre-processing. The corruptions are applied on various levels: local, global, object-level, and background (BG), and involve data addition, reduction, or alteration.}
         \vspace{-0.1cm}
         \label{tab:list}
         \resizebox{\textwidth}{!}{%
         \begin{tabular}{|c|c|c|c|c|c|c|c|c|c|c|c|c|}
      \hline
         ~ & \multicolumn{5}{c|}{Data Reduction} & \multicolumn{3}{c|}{Data Addition} & \multicolumn{4}{c|}{Data Alteration}      \\ \hline
Corruption & Global & Local & Object & BG & Part & Global & Local & Expand & Jitter & Local & BG & Incline \\\hline
sens. acc. & ~ & ~ & ~ & ~ & ~ & ~ & ~ & ~ & \checkmark & ~ & ~ & ~\\ \hline
sens. res. & \checkmark & ~ & ~ & ~ & ~ & \checkmark & ~ & ~ & ~ & ~ & ~ & ~\\ \hline
sens. loc. & ~ & \checkmark & ~ & ~ & \checkmark  & ~ & \checkmark & \checkmark & \checkmark & \checkmark & \checkmark & \checkmark\\ \hline
scene var. & ~ & ~ & \checkmark & \checkmark & \checkmark & ~ & ~ & \checkmark & ~ & ~ & ~ & \checkmark\\ \hline
preprocess & \checkmark & \checkmark & ~ & ~ & ~ & \checkmark & \checkmark & ~ & ~ & \checkmark & \checkmark & \checkmark\\ \hline
         \end{tabular}}
         \vspace{-0.3cm}
\end{table}

\noindent\textbf{Robustness Benchmark.}
We design point cloud corruptions to assess the robustness of point-based 3D object detectors. We analyze and benchmark 3D object detection architectures on the corrupted point clouds.

\noindent\textit{Corruption Types.}
We introduce corruptions to simulate variations that emanate from different sources, such as sensor accuracy/resolution/location, scene variations, and data pre-processing, as shown in Table \ref{tab:list}. We group corruptions into three main groups: (1) corruptions that lead to data reduction, (2) corruptions that lead to data addition, and (3) corruptions that alter the data. We apply data corruptions on various levels: local, global, object-level, and background.

\noindent\textit{Setup.} We introduce corruptions to ScanNetv2 \cite{scannet}, a richly annotated dataset of 3D meshes reconstructions of indoor scenes. It contains 1513 scenes out of which 312 form the validation set. We benchmark four different detectors on a corrupted validation set, For methods with multiple variants, we choose the top performing variant in our experiments.


    \noindent\textit{Evaluation Metrics.}
    We use Corruption Error (CE) and mean Corruption Error (mCE) \cite{benchmarking2019} for evaluation.
    We substitute top-1 error  $E_{s,c}^f$ in the original formula for mean Average Precision (mAP) to measure object detection performance. We report mAP at $0.25$ IoU threshold, and we choose VoteNet as our baseline. The Corruption Error is formulated as follows:
    \begin{align}
    \textrm{CE}_i = \frac{\sum_{l=1}^n (1 - \textrm{mAP}_{i,l})} {\sum_{l=1}^n (1 - \textrm{mAP}^{\textrm{VoteNet}}_{i,l})},
    \end{align}
     where $\textrm{mAP}_{i,l}$ is the mAP on a corrupted test set $i$ at corruption level $l$, $\textrm{mAP}^{\textrm{VoteNet}}_{i,l}$ is the baseline's mAP.
    mCE is the average of CE over $N$ corruptions:
    \begin{align}
    \textrm{mCE} = \frac{1}{N}\sum_{i=1}^N \textrm{CE}_i,
    \end{align}

    \subsection{Robustness Against Point Removal}
    While designing our suite of point removal corruptions, we use 6 different techniques in dropping points from a given scan: Drop Local, Drop Global, Drop Object, Drop Background, Drop Object Parts, and Drop Floor.
    
        \begin{figure}
            \vspace{-0.5cm}
             \centering
             \begin{subfigure}[b]{0.24\textwidth}
                 \centering
                 \includegraphics[width=\textwidth]{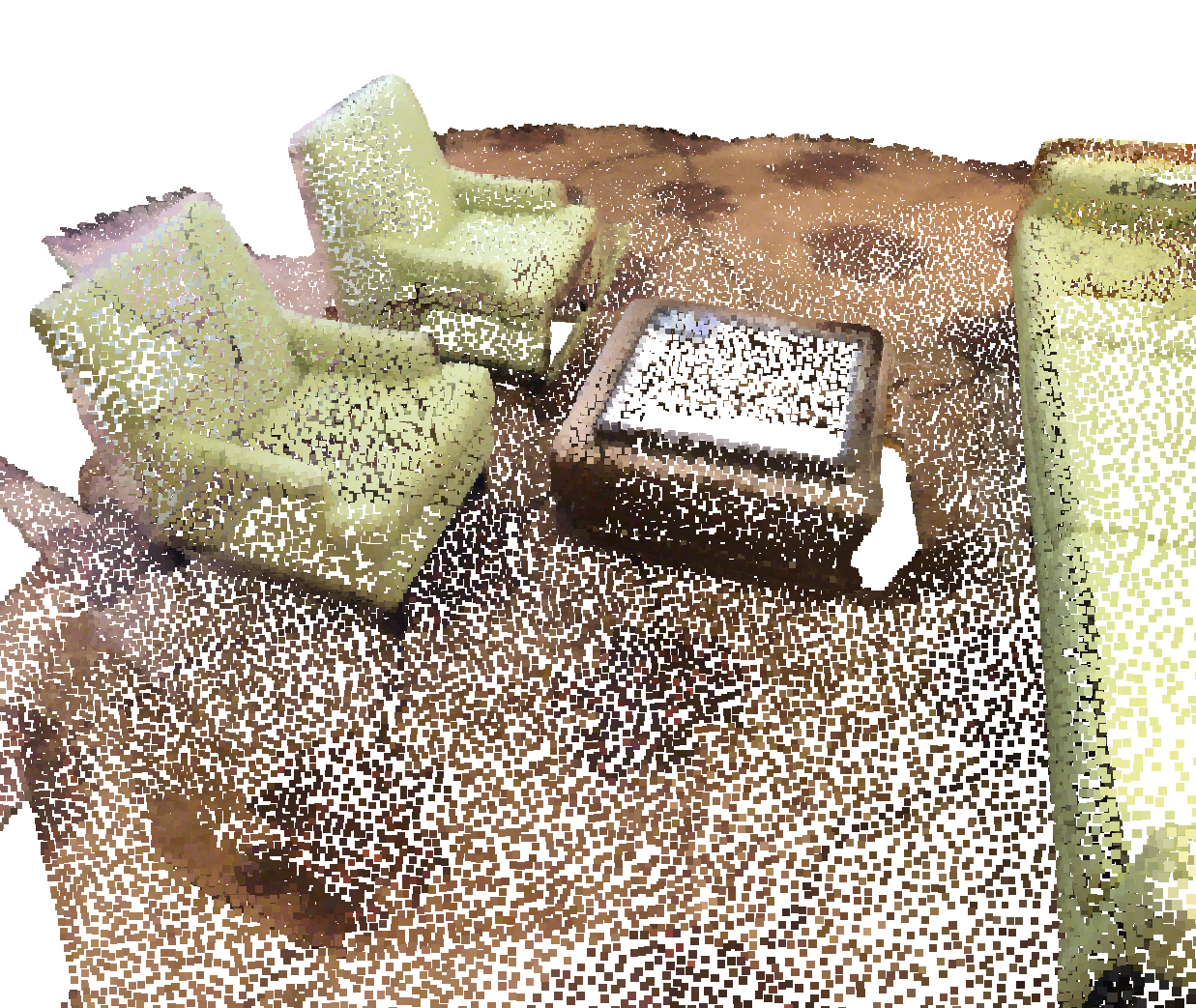}
                 \caption{Original Scene}
             \end{subfigure}
             \hfill
             \begin{subfigure}[b]{0.24\textwidth}
                 \centering
                 \includegraphics[width=\textwidth]{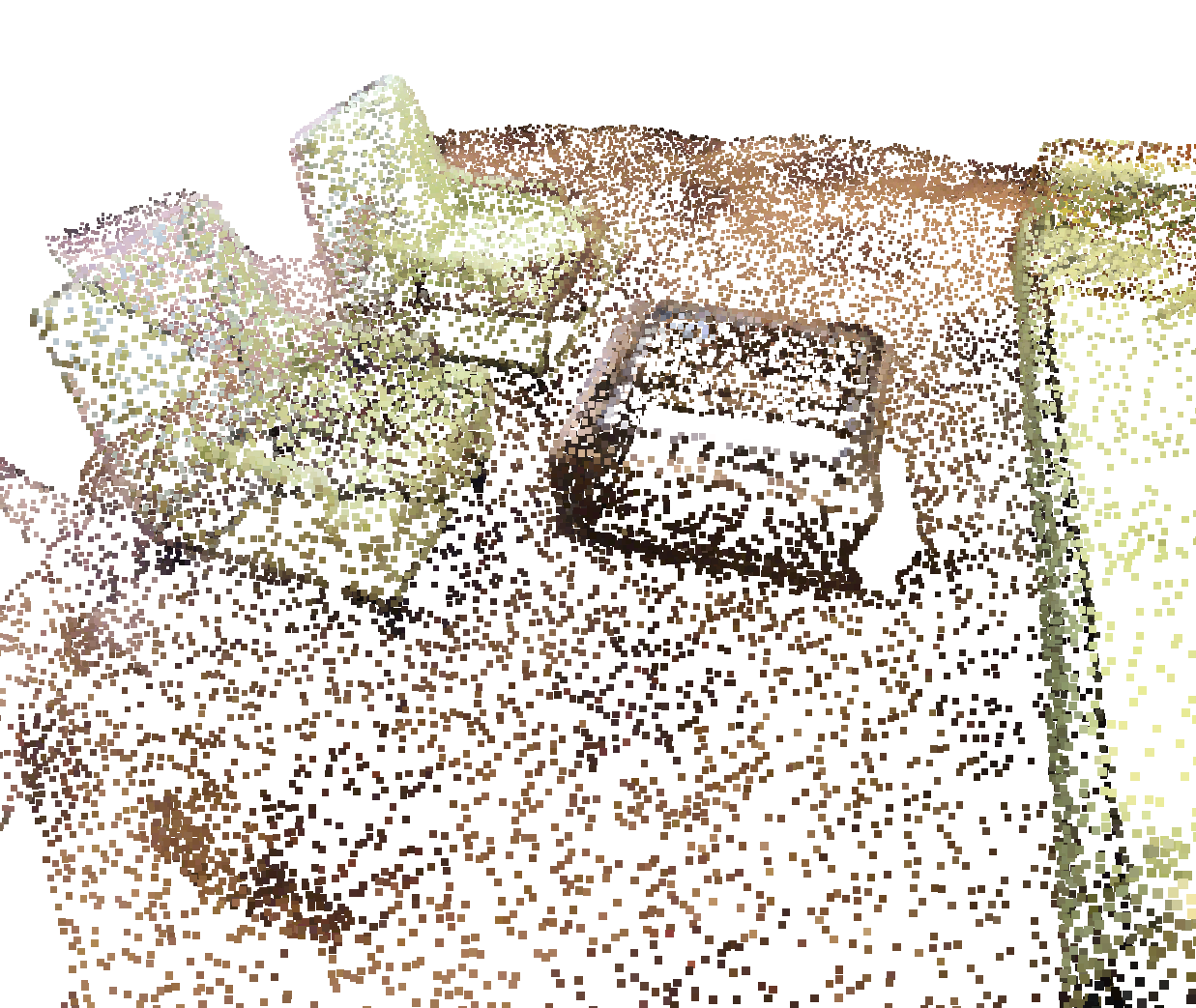}
                 \caption{Drop Global}
                 \label{fig:drop_global}
             \end{subfigure}
             \hfill
             \begin{subfigure}[b]{0.24\textwidth}
                 \centering
                 \includegraphics[width=\textwidth]{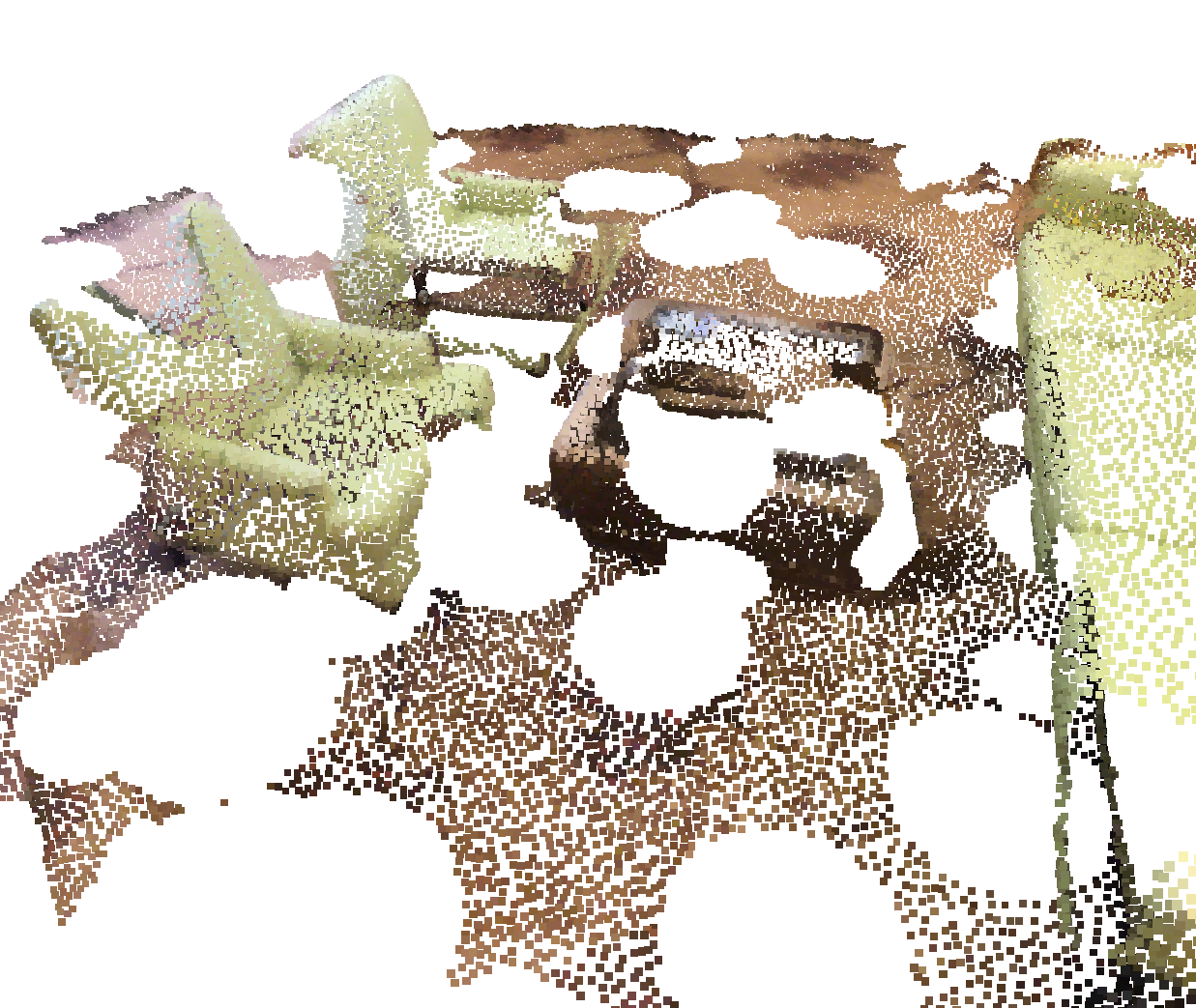}
                 \caption{Drop Local}
                 \label{fig:drop_local}
             \end{subfigure}
             \hfill
             \begin{subfigure}[b]{0.24\textwidth}
                 \centering
                 \includegraphics[width=\textwidth]{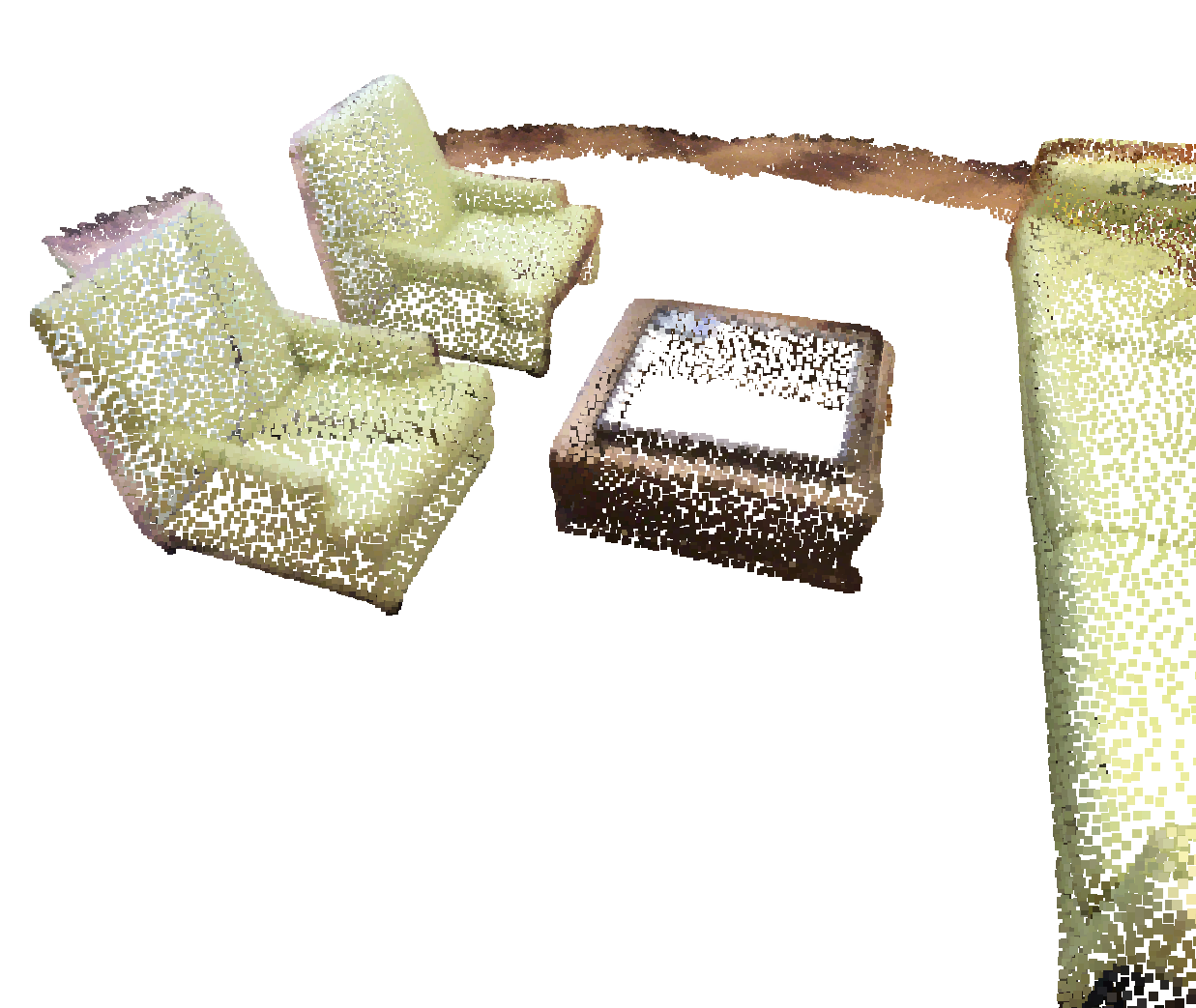}
                 \caption{Drop Floor}
                 \label{fig:drop_floor}
             \end{subfigure}\vspace{-0.1cm}
                \caption{An example scene with its corrupted versions (drop global, drop local, and drop floor). Removed points are replaced with duplicates of remaining points.}
                \label{fig:drop_global_local_floor}
                \vspace{-0.6cm}
        \end{figure}
    
        \subsubsection{Drop Global.}
        This corruption represents a uniformly distributed loss of points across a 3D scene, as shown in Fig.\ref{fig:drop_global}. We implement global drop at 5 severity levels, ranging from dropping 25\% of the points in the least severe level (level 1) to 75\% of the points in the most severe level (level 5). Dropped points are replaced by duplicates of remaining points to keep a fixed sized input. 
        \par
\begin{figure}
            \includegraphics[width=\textwidth]{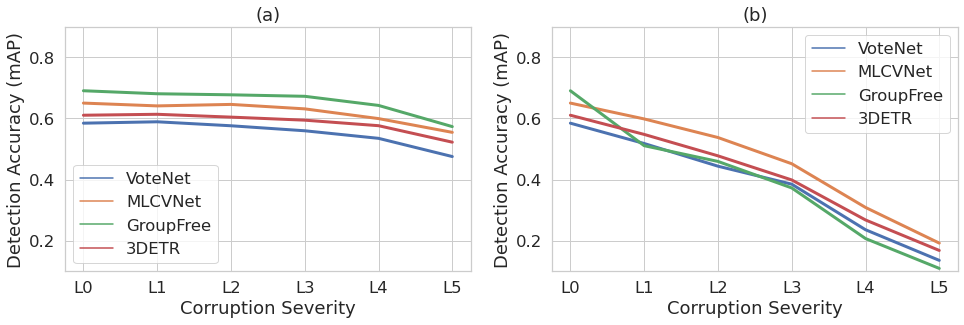} \vspace{-0.5cm}
            \caption{Detection Accuracy (mAP@0.25 IoU) for (a) Drop Global and (b) Drop Local corruptions over different architectures. The corrupted point clouds are evaluated on VoteNet \cite{votenet2019}, MLCVNet \cite{mlcvnet2020}, 3DETR \cite{3detr2021}, and GroupFree3D \cite{groupfree3d2021}.}\vspace{-0.7cm} \label{drop_both_graph}
        \end{figure}

        We report detection accuracy with global drop in Fig. \ref{drop_both_graph}a. All detectors show robustness against this corruption with mAP decrease of less than 2.5\% at level 3 (50\% of the points dropped) and decent performance at higher levels.

        \subsubsection{Drop Local.}
        In contrast to drop global, drop local represents a clustered loss of points, similar to the occlusion effect (Fig.\ref{fig:drop_local}). In each scan, we use a uniform random distribution function to determine the position, size, and the number of clusters. We implement drop local in 5 increasing severity levels. In the least severe corruption (level 1), 25\% of all points are uniformly selected in clusters and replaced by duplicates of the remaining points. With increasing the level of severity, the dropout ratio is increased by 12.5\% up to level 5 where 75\% of the points are dropped.
        \par
        Fig. \ref{drop_both_graph}b demonstrates mAP@0.25 values across different drop local levels. At the first level, VoteNet, MLCVNet, and 3DETR's drop rate does not exceed 7\% while Group-Free suffers greatly, with $3x$ reduction compared to other methods. We reason that this significant drop is due to Group-Free's reliance on local patch and object information, which is heavily affected by the local drop.
        

        \subsubsection{Drop Floor.}
        Drop Floor represents the loss of all the points that make up the floor in the 3D scene (Fig.\ref{fig:drop_floor}). This aims at analyzing the ability of different methods to detect objects from shape information, without relying on cues from the floor plane. In each 3D scan, we drop all the points occurring below the 1st percentile of the points along the upward direction. 
        
        \begin{wraptable}{r}{4cm}
        \centering
        \small
        \vspace{-8mm}
         \caption{Detectors mAP before and after applying Drop Floor corruption.}
         \vspace{-2mm}
         \label{tab:drop-floor}
         \begin{tabular}{|l|c|c|}
         \hline
         & \multicolumn{2}{c|}{mAP} \\
         \makecell{Method}  & \makecell{Before} & \makecell{After} \\
         \hline
         VoteNet & 58.44 & 43.81 \\
         MLCVNet & 65.01 & 52.80 \\
         Group-Free &69.05& 63.86 \\
         3DETR &  61.04 & 55.21 \\
         \hline
         \end{tabular}
         \vspace{-0.6cm}
         \end{wraptable}
         
        \par
        The floor plane in indoor scans provides a good contextual cue, and the estimated floor height can be used as an additional input feature. The floor height is usually estimated using a percentile along the vertical dimension. Since natural scenes are very diverse, this estimation is not always correct. We show our results in Table \ref{tab:drop-floor}. We observe that Group-Free and 3DETR are more robust than VoteNet and MLCVNet. This is due to the latter methods using estimated height as an additional input feature. 
        
        \begin{figure}
             \centering
             \vspace{-0.5cm}
             \begin{subfigure}[b]{0.24\textwidth}
                 \centering
                 \includegraphics[width=\textwidth]{corrupted_point_clouds/original_312_00.png}
                 \caption{Original Scene}
             \end{subfigure}
             \hfill
             \begin{subfigure}[b]{0.24\textwidth}
                 \centering
                 \includegraphics[width=\textwidth]{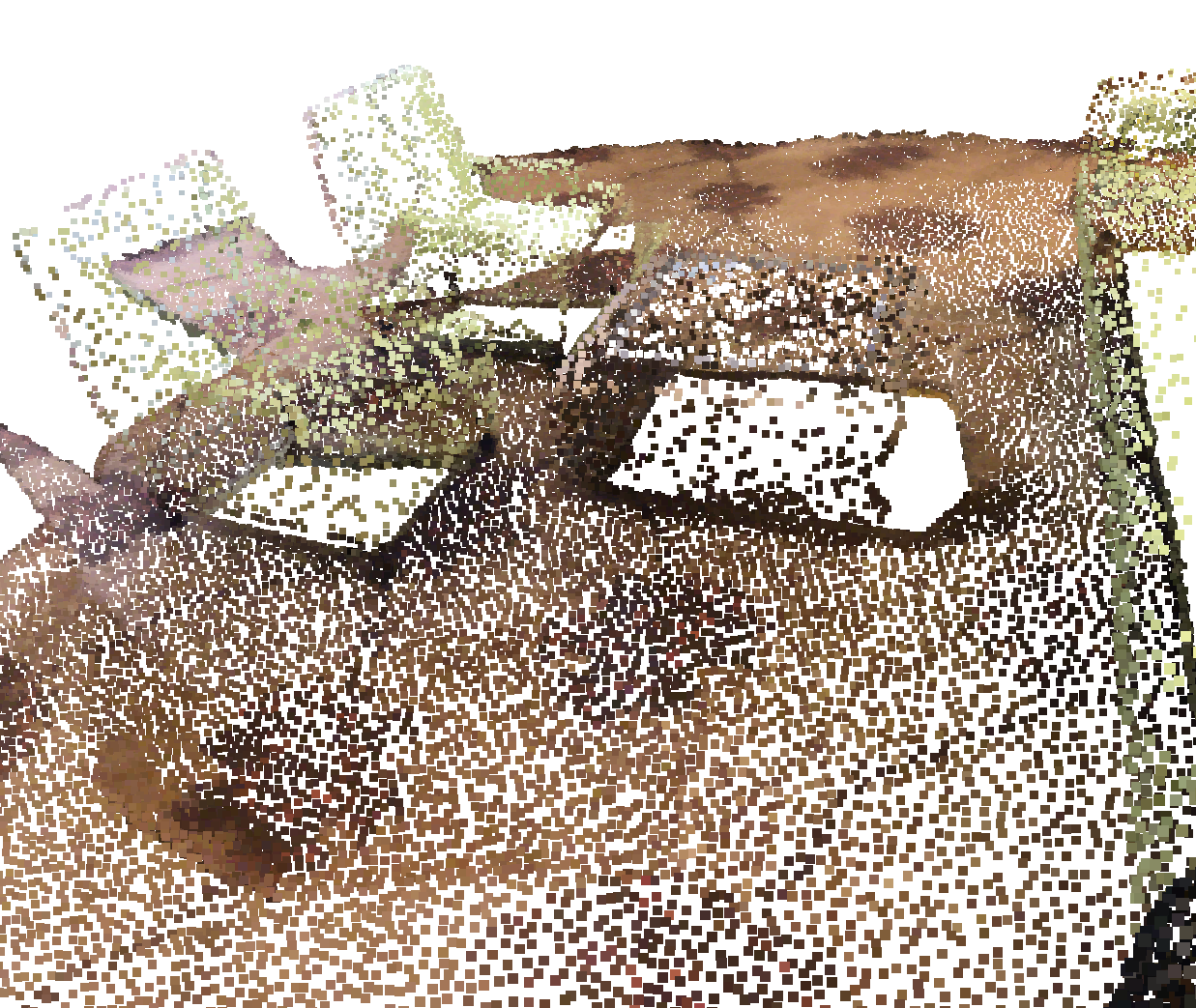}
                 \caption{Drop Object}
                 \label{fig:drop_object}
             \end{subfigure}
             \hfill
             \begin{subfigure}[b]{0.24\textwidth}
                 \centering
                 \includegraphics[width=\textwidth]{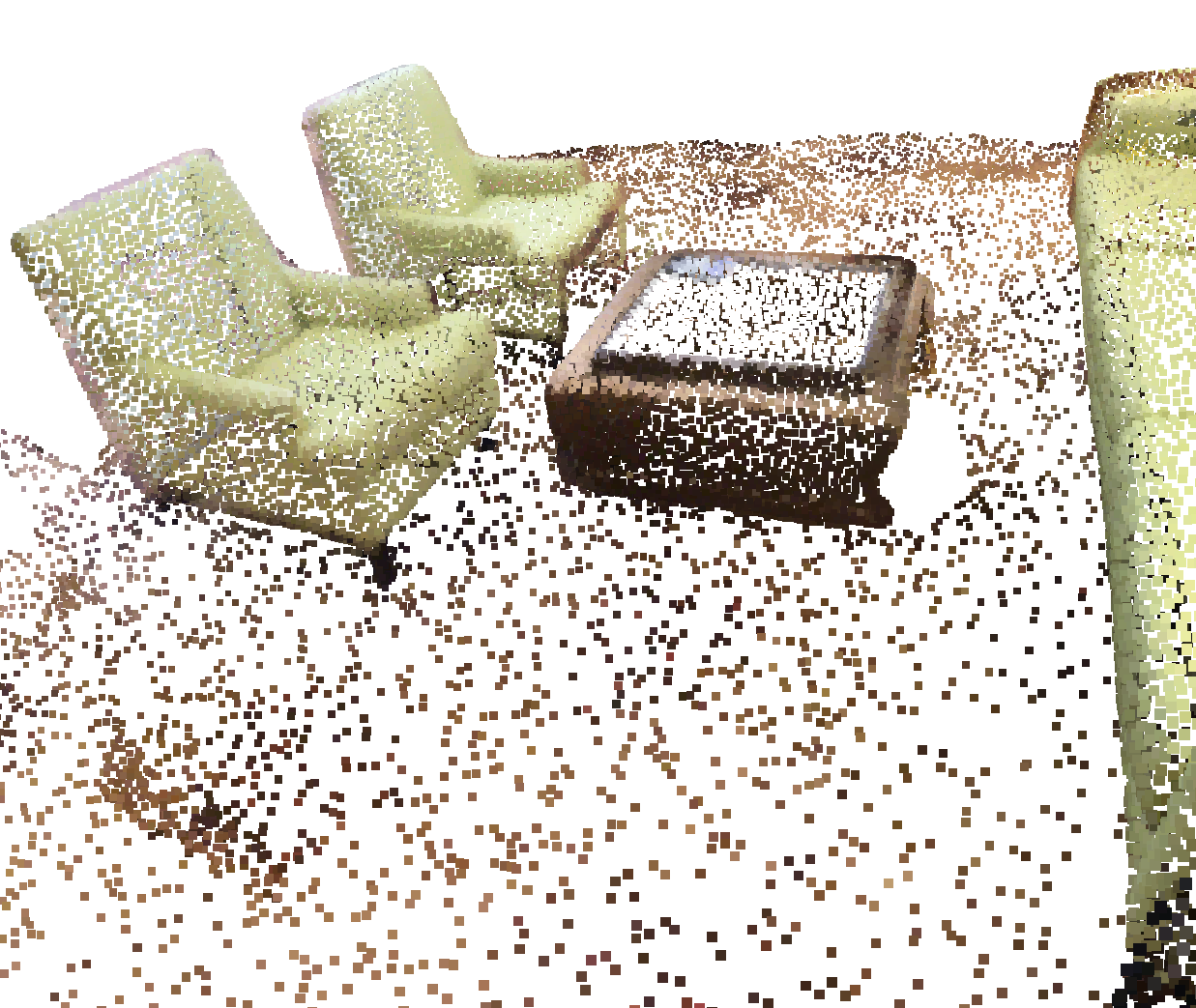}
                 \caption{Drop Bckgrnd}
                 \label{fig:drop_Bckgrnd}
             \end{subfigure}
             \hfill
             \begin{subfigure}[b]{0.24\textwidth}
                 \centering
                 \includegraphics[width=\textwidth]{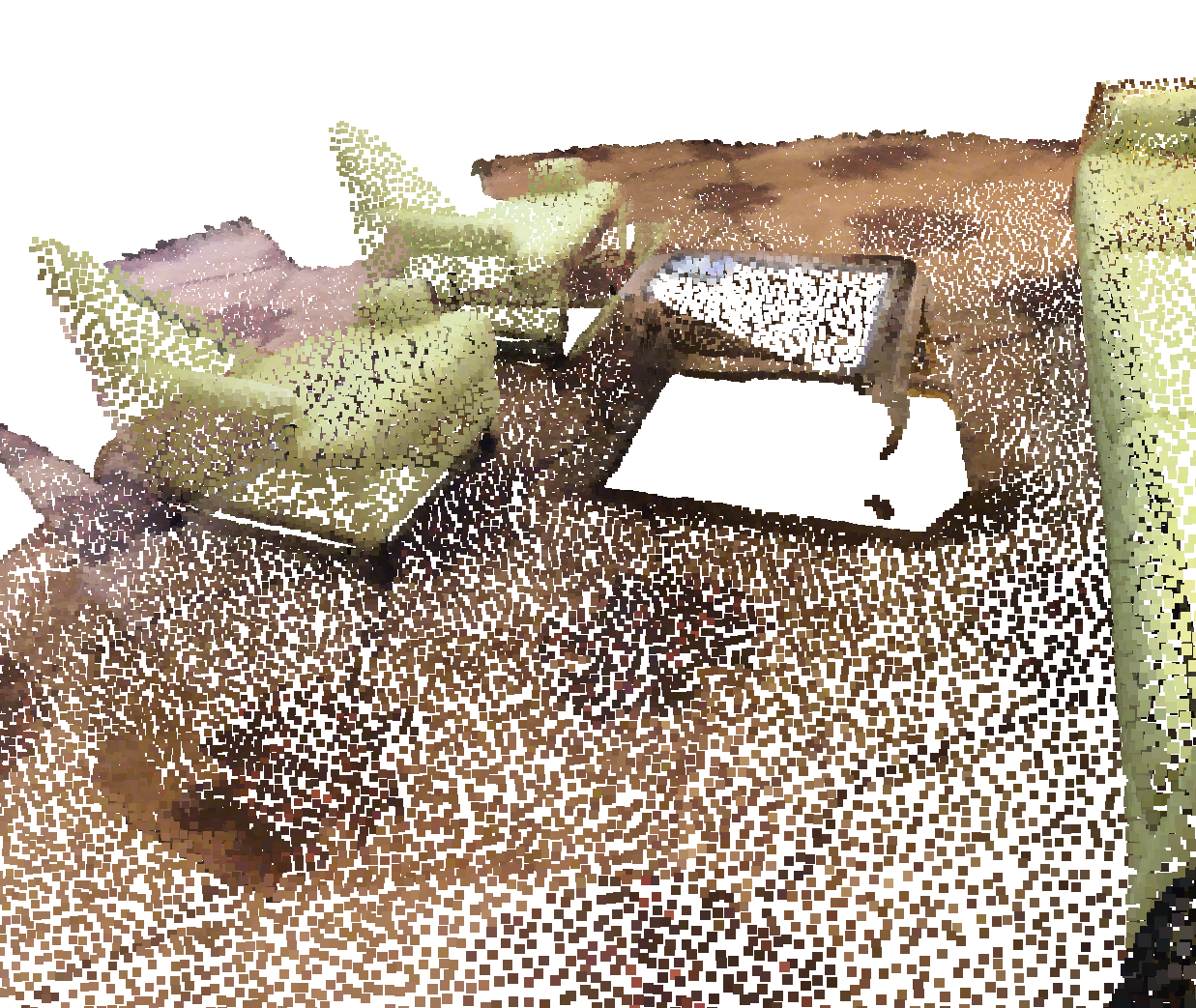}
                 \caption{Drop Obj Parts}
                 \label{fig:drop_amodal}
             \end{subfigure}\vspace{-0.1cm}
                \caption{An example scene with its corrupted versions (Drop Object, Drop Background, and Drop Object Parts). Removed points are replaced with duplicates of the remaining points. Each corruption is implemented in 5 levels of severity where the ratio of the dropped points increases with every level.}\vspace{-0.9cm}
                \label{fig:drop_object_bg_amodal}
        \end{figure}
        
        \subsubsection{Drop Object.}
        Similar to drop global, in this corruption, we drop random points uniformly. However, the selected points are taken only from objects of interest excluding the background, as shown in Fig. \ref{fig:drop_object}. We implement this with 5 gradually increasing levels of severity. In level 1, 25\% of the object points are dropped and replaced by duplicates of the remaining points from the same object. At the highest severity level, 75\% of the object points are dropped. 
        \par
\begin{figure}
            \includegraphics[width=\textwidth]{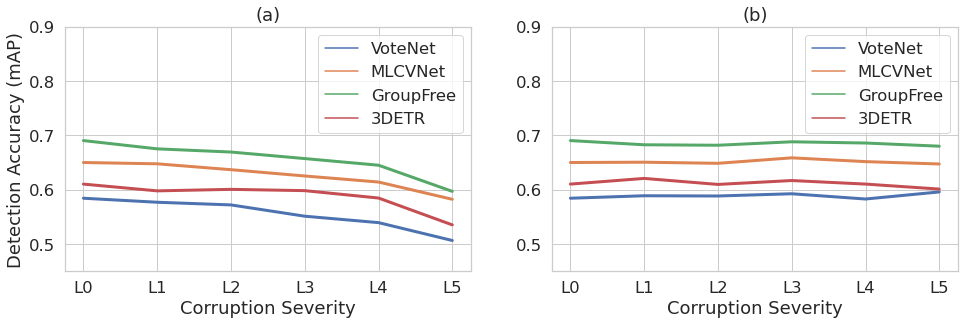}\vspace{-0.5cm}
            \caption{Detection Accuracy (mAP@0.25 IoU) for (a) Drop Object and (b) Drop Background corruptions over different architectures.}\vspace{-1cm} \label{drop_object_both_graph}
        \end{figure}

        Results are shown in Fig. \ref{drop_object_both_graph}a. We note that the performance of all models steadily declines as the severity level increases. It is noteworthy to mention that even with 62.5\% of the object's points being dropped, all detectors were able to perform exceptionally well with only a maximum drop of around 3\% mAP. This suggests that detectors rely on contextual cues to infer the object from surrounding points. Pushing the severity level up to 75\% drop rate leads to performance degradation due to the high loss in local object shape information. 

        \subsubsection{Drop Background.}
        We also consider scenarios where the missing points are exclusively background points, which belong to the floor, wall, or the ceiling. An example of this corruption is shown in Fig.\ref{fig:drop_Bckgrnd}. We also implement this corruption with 5 levels of severity, with 25\% to 75\% drop in background points drop. 
        \par
        Do detectors rely on local shape information to infer the object class and position or rely more on context? In this corruption, we aim at analyzing the behavior of detectors once context information is gradually dropped. Results are shown in Fig. \ref{drop_object_both_graph}b. All methods show robustness against removing up to 75\% of the background points, with less that 1\% decrease. This shows that minor scene structure is helpful to encode contextual information. 

        \subsubsection{Drop Object Parts.}
        Amodal object detection aims at estimating the physical size and structure of an object even if partially visible, and has benefits in applications such as robotics navigation, grasp estimation, and Augmented Reality. We investigate the ability of 3D detectors to estimate amodal bounding boxes given partial objects. For all object instances, we slice a portion of the object on a random axis and a random direction, where the portion size increases from 10\% to 50\% of the object. An example of this corruption is shown in Fig.\ref{fig:drop_amodal}.
        \par
\begin{figure}
            \includegraphics[width=\textwidth]{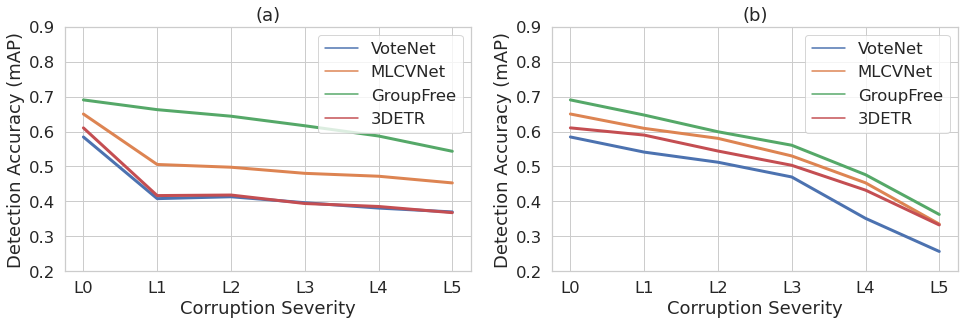}
            \vspace{-0.4cm}
            \caption{Detection Accuracy (mAP@0.25 IoU) for (a) Drop Object Parts and (b) Drop Object Parts with updated bboxes over different architectures.}
            \label{drop_amodalbbox_graph}
        \end{figure}

        We report detection results using the original ground-truth in Fig. \ref{drop_amodalbbox_graph}a. With minor corruption, Group-Free was able to maintain good performance compared to the other methods. We believe that this is facilitated by the global sharing of shape, size, and location information in the self-attention module, allowing detection with a broader understanding. We also show evaluation with updated bounding boxes that enclose the remaining object part in Fig.\ref{drop_amodalbbox_graph}a. This shows that methods tend to regress to the smaller bounding box at low corruption levels, but favor the amodal bounding box at high corruption levels. 
        
        \begin{table}[htp]
        \vspace{-1.2cm}
        \centering
        \caption{Corruption Error(CE) and mean Corruption Error(mCE) across Points Removal corruptions in all architectures.}\vspace{-0.2cm}
        \label{tab:CE-PointsRemoval}
        \resizebox{\textwidth}{!}{%
        \begin{tabular}{l|c|c|c|c|c|c|c|c}
        \hline
         Architectures & mAP $\uparrow$ & mCE $\downarrow$ & Global &         Local & Object & Background & Object-parts & Floor  \\
        \hline
        VoteNet\cite{votenet2019} & 58.44 & 1.000 & 1.000 &1.000 &1.000 &1.000 & 1.000 &1.000          \\
        MLCVNet\cite{mlcvnet2020} & 65.01 & 0.897 & 0.851 & 0.887 & 0.840 & 0.850 & 0.855 & 0.868 \\
        Group-Free\cite{groupfree3d2021} & 69.05 & 0.842 & 0.774  &	1.018 &	0.779  &	0.770  &	0.642  &	0.685  \\
        3DETR\cite{3detr2021} & 61.04 & 1.067 & 0.922	& 0.957	& 0.924	& 0.946	& 0.995 &	0.832 \\
        \hline
        \end{tabular}}\vspace{-0.3cm}
        \end{table}

We show Corruption Error(CE) and mean Corruption Error(mCE) for all point removal corruptions in Table \ref{tab:CE-PointsRemoval}. We observe that the Transformer-based Group-free method is the most robust against point removal corruptions. The least robust is also a Transformer-based method (3DETR). This suggests that employing Transformers at the patch and object level realizes more robustness than extracting point-level features used in 3DETR.  

    \subsection{Robustness Against Point Addition}
    During the acquisition process of point cloud scenes, outlier points and noise may be captured as well. 
    We introduce 3 corruptions: Add Global, Add Local and Add Below Floor. Each corruption aims at understanding how models deal with such variations. We apply the corruption with different levels of severity to model different scenarios in the real-world setting.
    
        \begin{figure}[t!]
             \centering
             \begin{subfigure}[b]{0.24\textwidth}
                 \centering
                 \includegraphics[width=\textwidth]{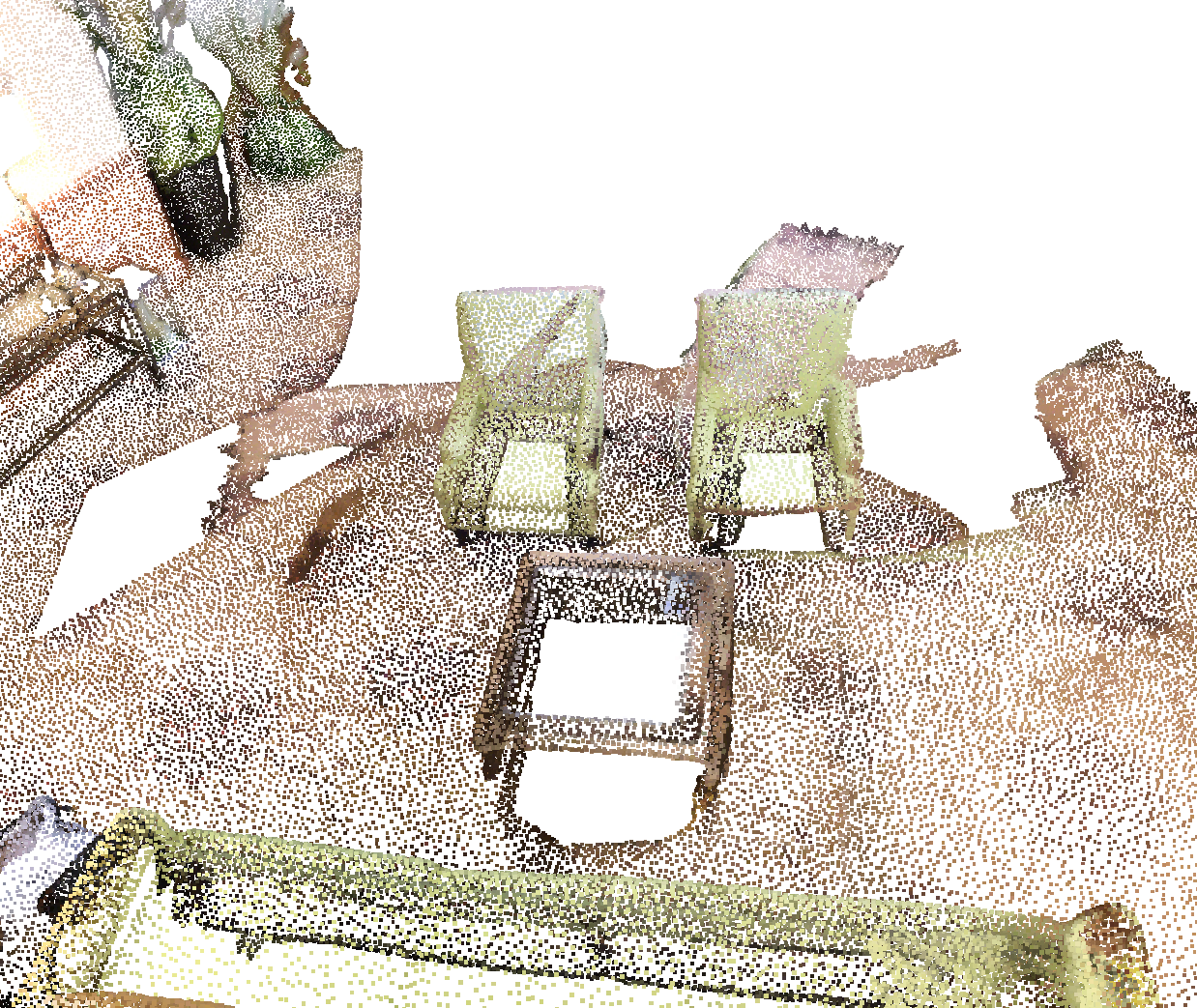}
                 \caption{Original Scene}
             \end{subfigure}
             \hfill
             \begin{subfigure}[b]{0.24\textwidth}
                 \centering
                 \includegraphics[width=\textwidth]{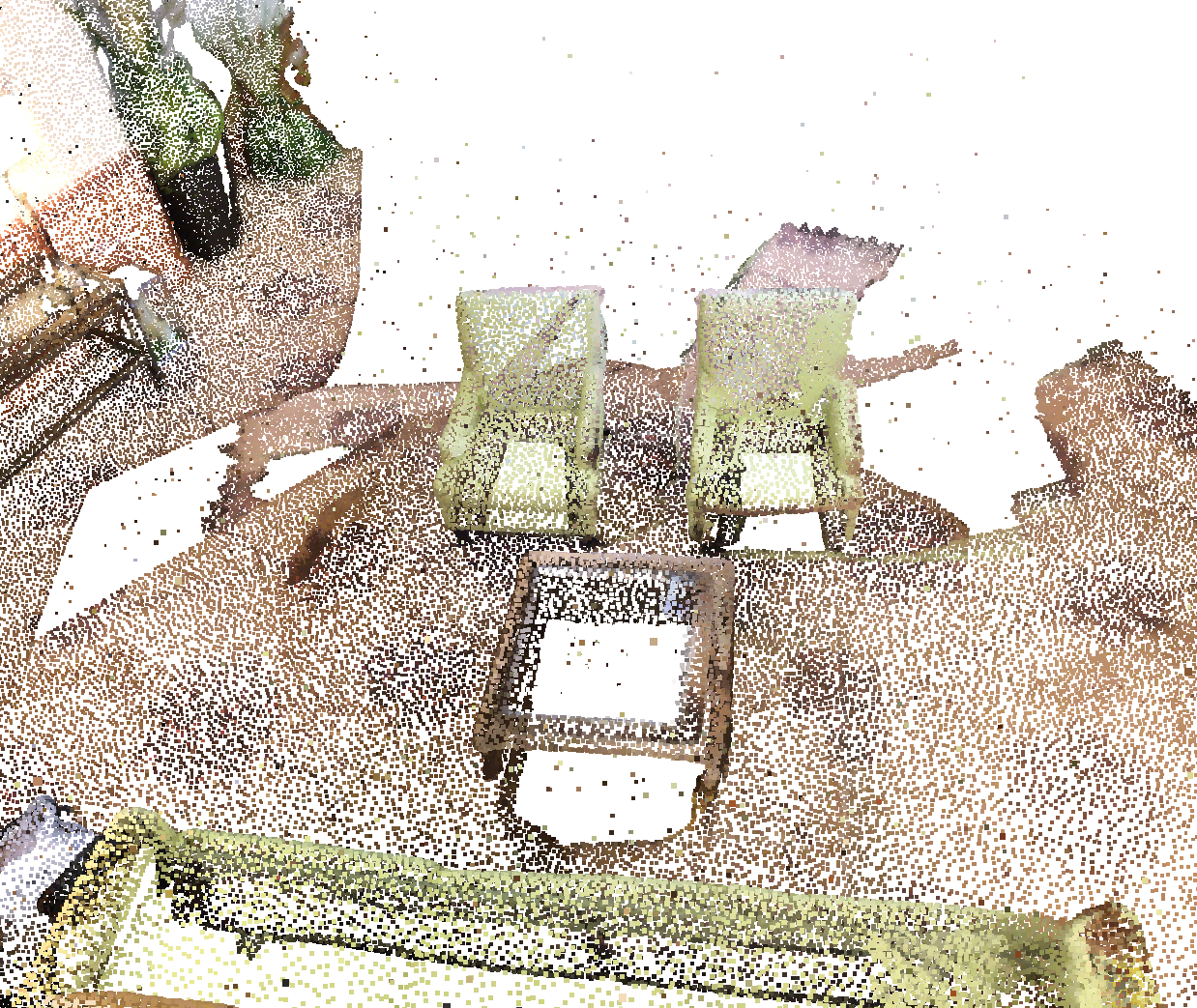}
                 \caption{Add Local}
                 \label{fig:add_local}
             \end{subfigure}
             \hfill
             \begin{subfigure}[b]{0.24\textwidth}
                 \centering
                 \includegraphics[width=\textwidth]{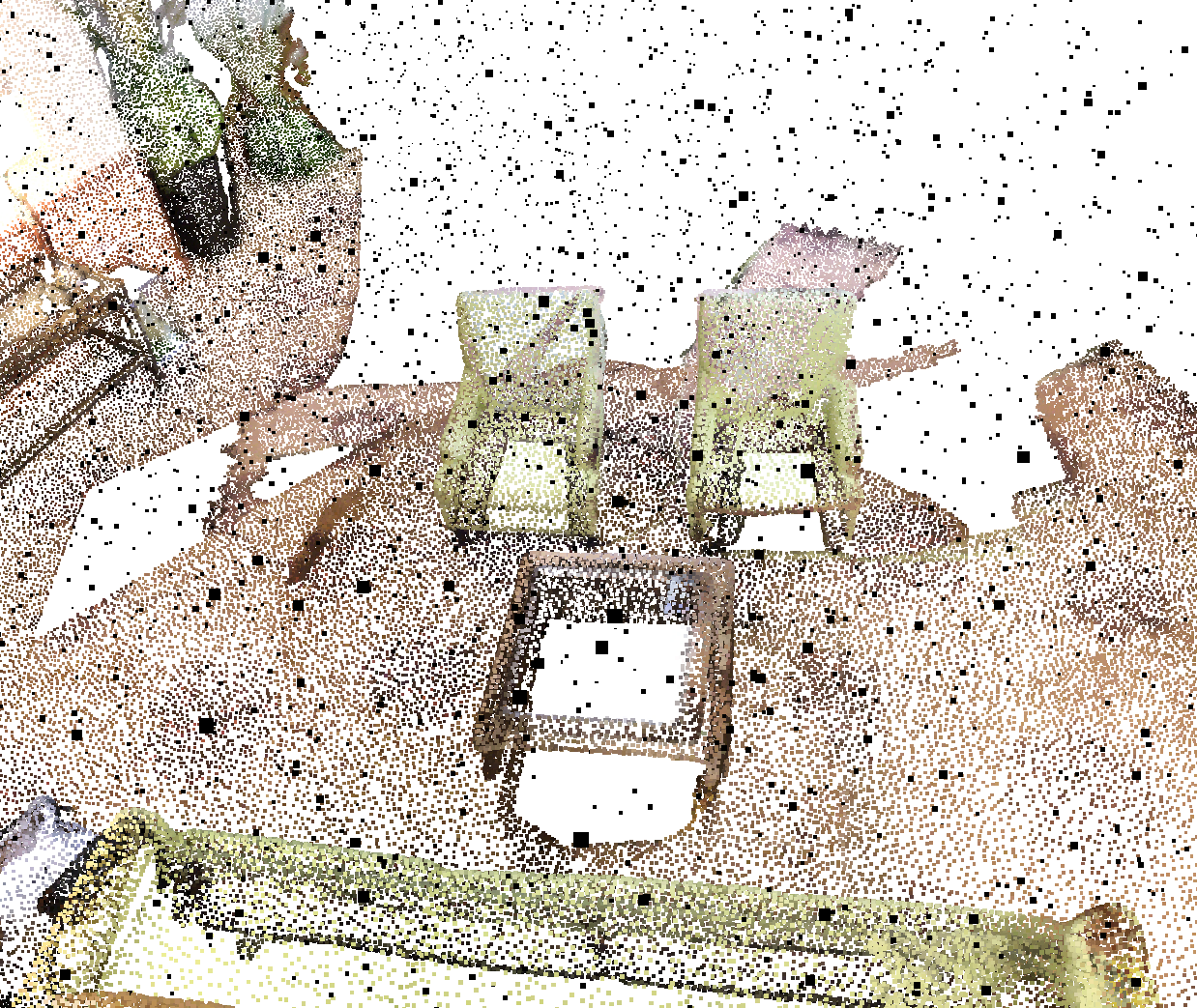}
                 \caption{Add Global}
                 \label{fig:add_global}
             \end{subfigure}
             \hfill
             \begin{subfigure}[b]{0.24\textwidth}
                 \centering
                 \includegraphics[width=\textwidth]{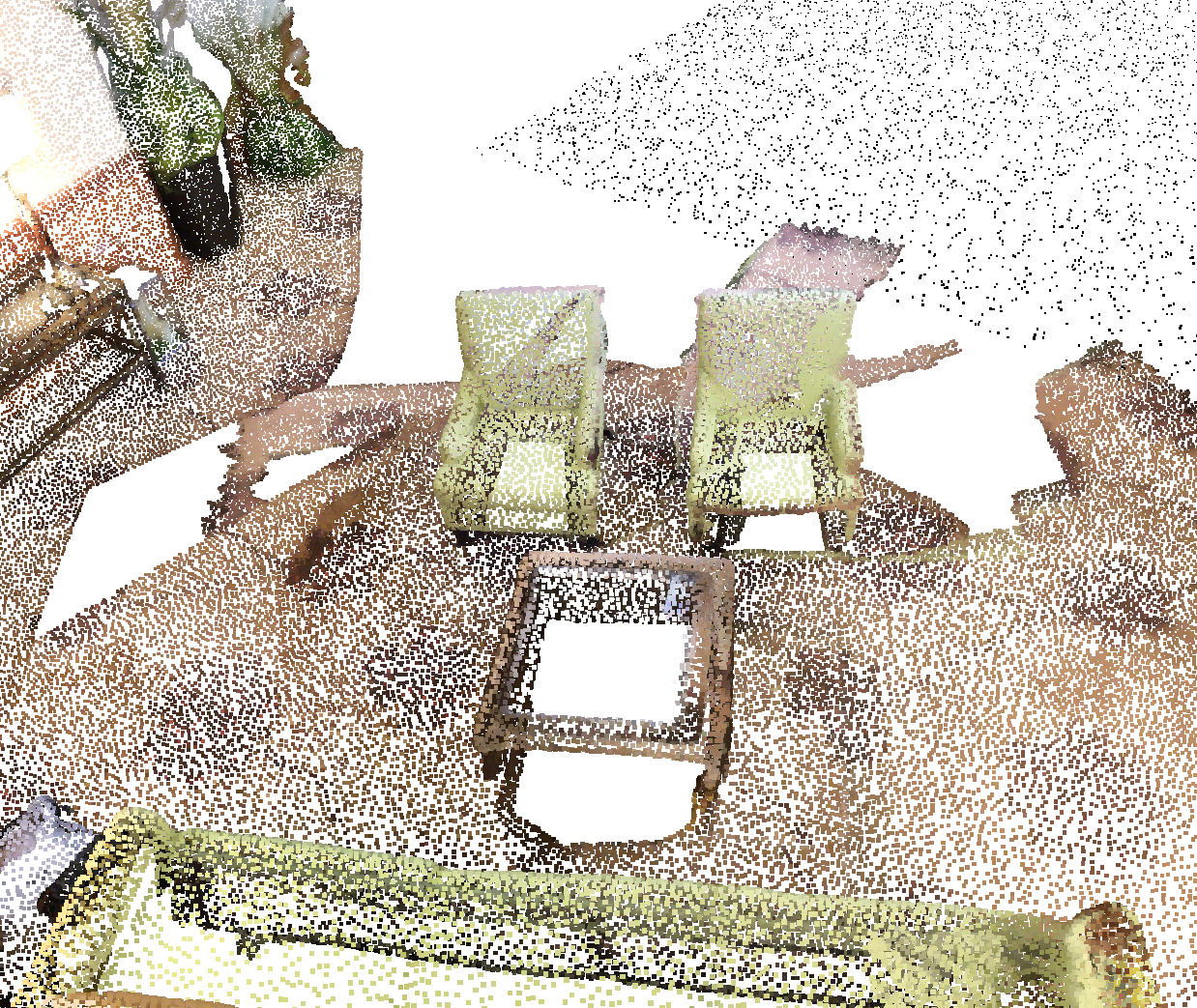}
                 \caption{Expand Scene}
                 \label{fig:add_floor}
             \end{subfigure}\vspace{-0.2cm}
                \caption{An example scene with its corrupted versions (add local, add global, and expand scene). The introduced points are given no labels and no colors. Before adding points, the original scene is down-sampled proportionally to ensure the final scene has the initial number of points. Each corruption is implemented in 5 levels of severity where the ratio of the added points increases with every level.}\vspace{-0.3cm}
                \label{fig:add_global_local_floor}
        \end{figure}
        
        \begin{figure}
            \includegraphics[width=\textwidth]{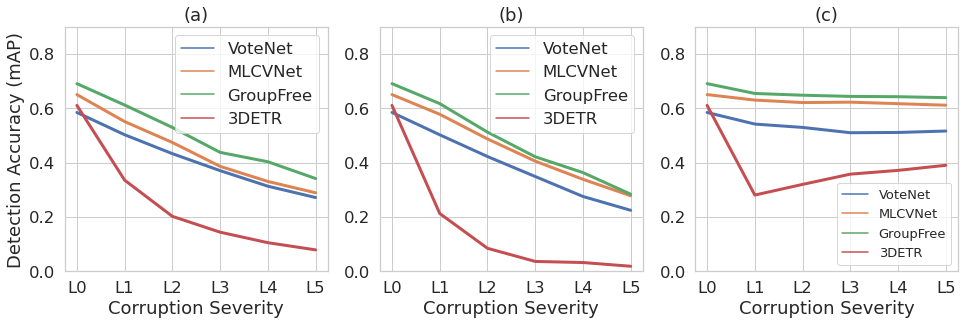}\vspace{-0.2cm}
            \caption{Detection Accuracy (mAP@0.25 IoU) for (a) Add Global, (b) Add Local, and (c) Scene Expansion corruptions over different architectures.} \vspace{-0.5cm}\label{add_both_expand_graph}
        \end{figure}
    
        \subsubsection{Add Global.}
        We randomly sample (X,Y,Z) components from a uniform distribution to generate points that fall within the range of the given scene. We add \{1,2,3,4,5\}\% of the scene’s total number of points for the five levels (Fig.\ref{fig:add_global}).
        \par
        We show results in Fig. \ref{add_both_expand_graph}a. With the addition of 1\% of points, which is only 400 points, the performance of all detectors drops substantially. Specifically, Group-Free, VoteNet, MLCVNet, and 3DETR drop in accuracy about 8\%, 8\% 10\%, and 28\% respectively. We speculate that the reason behind the huge drop in 3DETR performance is the global view of the scene captured by self-attention modules. In the feature learning stage, every point is affected by all the other points due to the self-attention mechanism. This self-attention will include the noise as well leading to poor point feature representation. 
        
        \subsubsection{Add Local.}
        We follow the setting of ModelNet40-C \cite{benchmarking2022}. First, we normalize the scene and then randomly select C centroids, where C $\in [1, 312]$. Around each centroid, we generate a random number of points. We draw the neighboring point coordinates from a Normal distribution $\mathcal{N}(\boldsymbol{\mu}_i, \sigma_i^2\boldsymbol{\textrm{I}})$, where $\boldsymbol{\mu}_i$ is the i-th centroid coordinate and $\sigma_i\in \mathcal{U}(0.075, 0.125)$. The number of points added ranges from 1\% to 5\% of the total number of points. An example is shown in Fig.\ref{fig:add_local}.
        \par
        In Fig. \ref{add_both_expand_graph}b, we show Add Local corruptions results. We note that 3DETR performance deteriorates the most followed by Group-Free. The latter is due to the self-attention module that models the interaction between object features. The added noise is at the local level which affects local interactions.

        \subsubsection{Scene Expansion.} 
        This corruption aims to replicate a scene with multiple floor planes like a stairwell for instance. We introduce new points in the form of a plane that is below the original floor, but outside the scene (Fig.\ref{fig:add_floor}). 
        Fig. \ref{add_both_expand_graph}c shows that point-based architectures tend to be relatively robust to such corruptions. On the other hand, the performance of the transformer-based architecture drops significantly at the lowest level of severity.

    \begin{table}[t!]
        \centering
        \caption{Corruption Error(CE) and mean Corruption Error(mCE) across Point Addition corruptions in all architectures.}\vspace{-0.2cm}
        \label{tab:CE-PointsAddition}
        \begin{tabular}{l|c|c|c|c|c}
        \hline
         Architectures & mAP $\uparrow$ & mCE $\downarrow$ & Global &         Local & Scene-Expansion   \\
        \hline
        VoteNet\cite{votenet2019} & 58.44 & 1.000 & 1.000 &1.000 &1.000  \\
        MLCVNet\cite{mlcvnet2020} & 65.01 & 0.897 & 0.955 & 0.903 & 0.794 \\
        Group-Free\cite{groupfree3d2021} & 69.05 & 0.842 & 0.861 &0.869 &	0.741 \\
        3DETR\cite{3detr2021} & 61.04 & 1.067 & 1.329 &1.430 &1.372 \\
        \hline
        \end{tabular}\vspace{-0.3cm}
        \end{table}
    
    
    As illustrated in Table \ref{tab:CE-PointsAddition}, we report CE and mCE over Points Addition corruptions. 3DETR performs the worst in all three corruptions. The Detector scores the largest CE of $1.430$ in Add Local corruption. We conclude that while the transformers' wide receptive field is desirable, the self-attention mechanism that is needed to achieve that is extremely sensitive to noise.
    
    \subsection{Robustness Against Alterations}
    Sensors noise leads to spatial inaccuracies, where points' position, scale, and angle are wrongly measured. We aim to capture the same kind of operation and effect in the subsequent corruptions. We introduce Jitter, Local Noise, Background Noise, and Floor Plane Inclination.  
    
        \begin{figure}[t]
             \centering
             \begin{subfigure}[b]{0.24\textwidth}
                 \centering
                 \includegraphics[width=\textwidth]{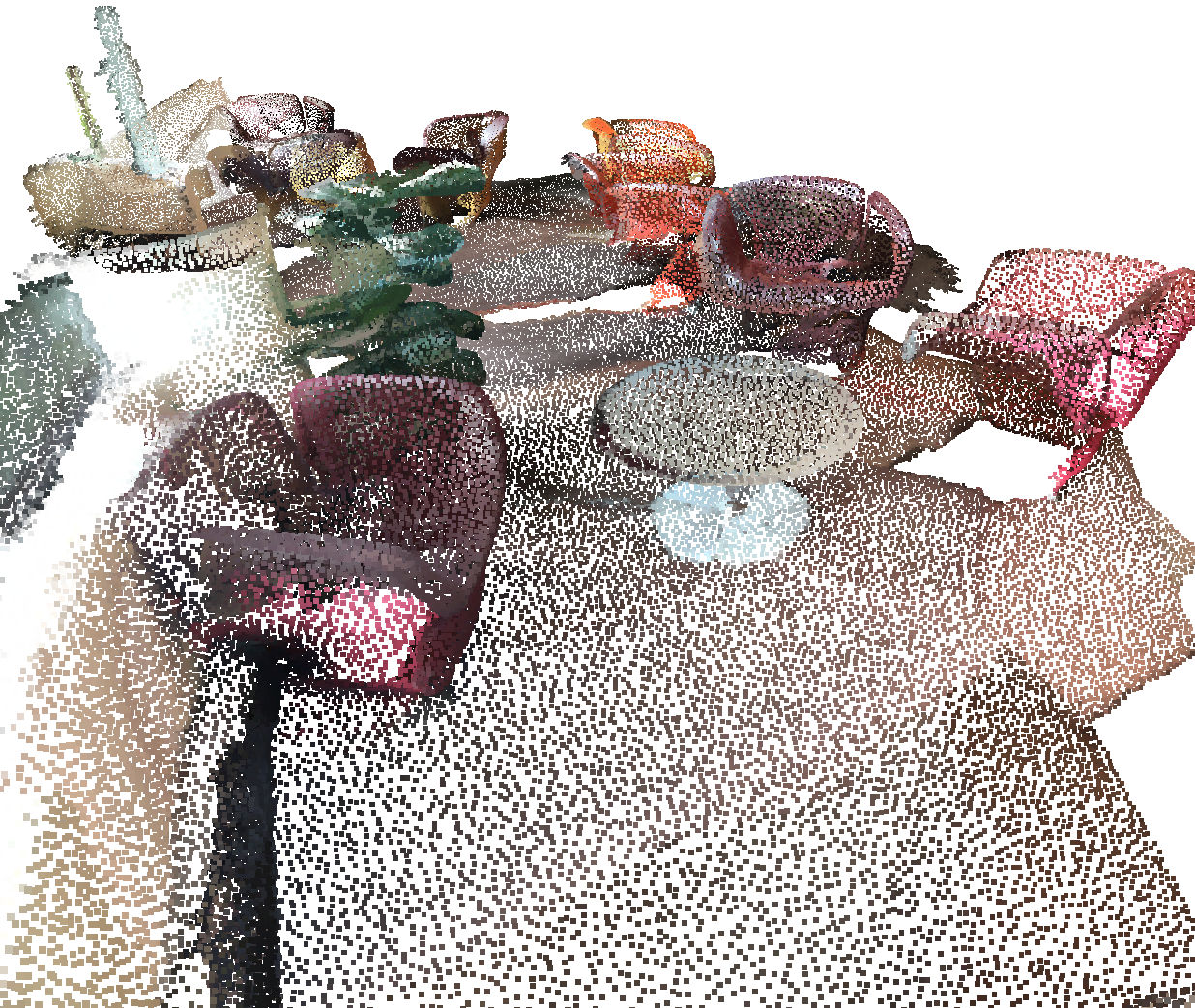}
                 \caption{Original Scene}
             \end{subfigure}
             \hfill
             \begin{subfigure}[b]{0.24\textwidth}
                 \centering
                 \includegraphics[width=\textwidth]{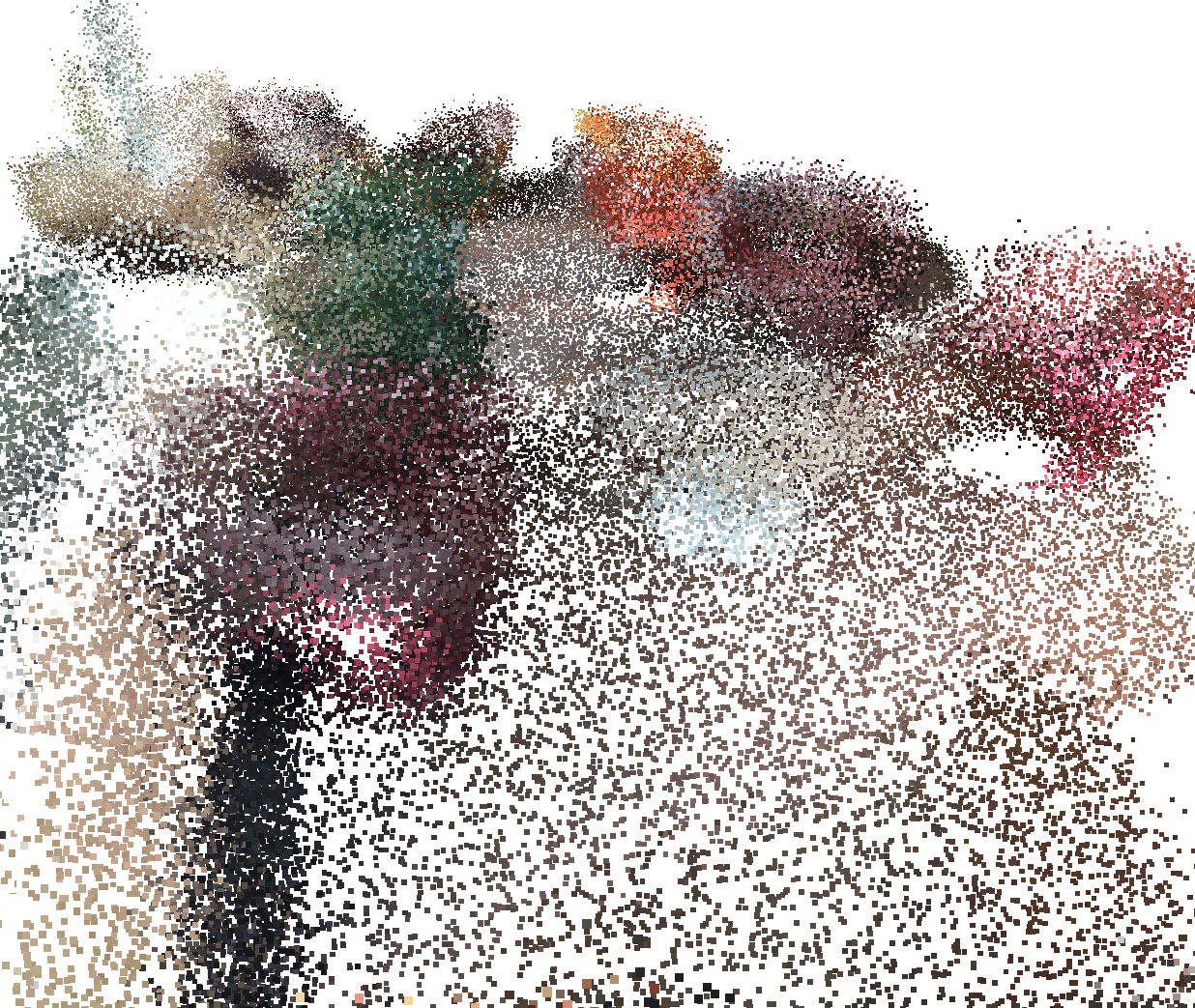}
                 \caption{Jitter}
             \end{subfigure}
             \hfill
             \begin{subfigure}[b]{0.24\textwidth}
                 \centering
                 \includegraphics[width=\textwidth]{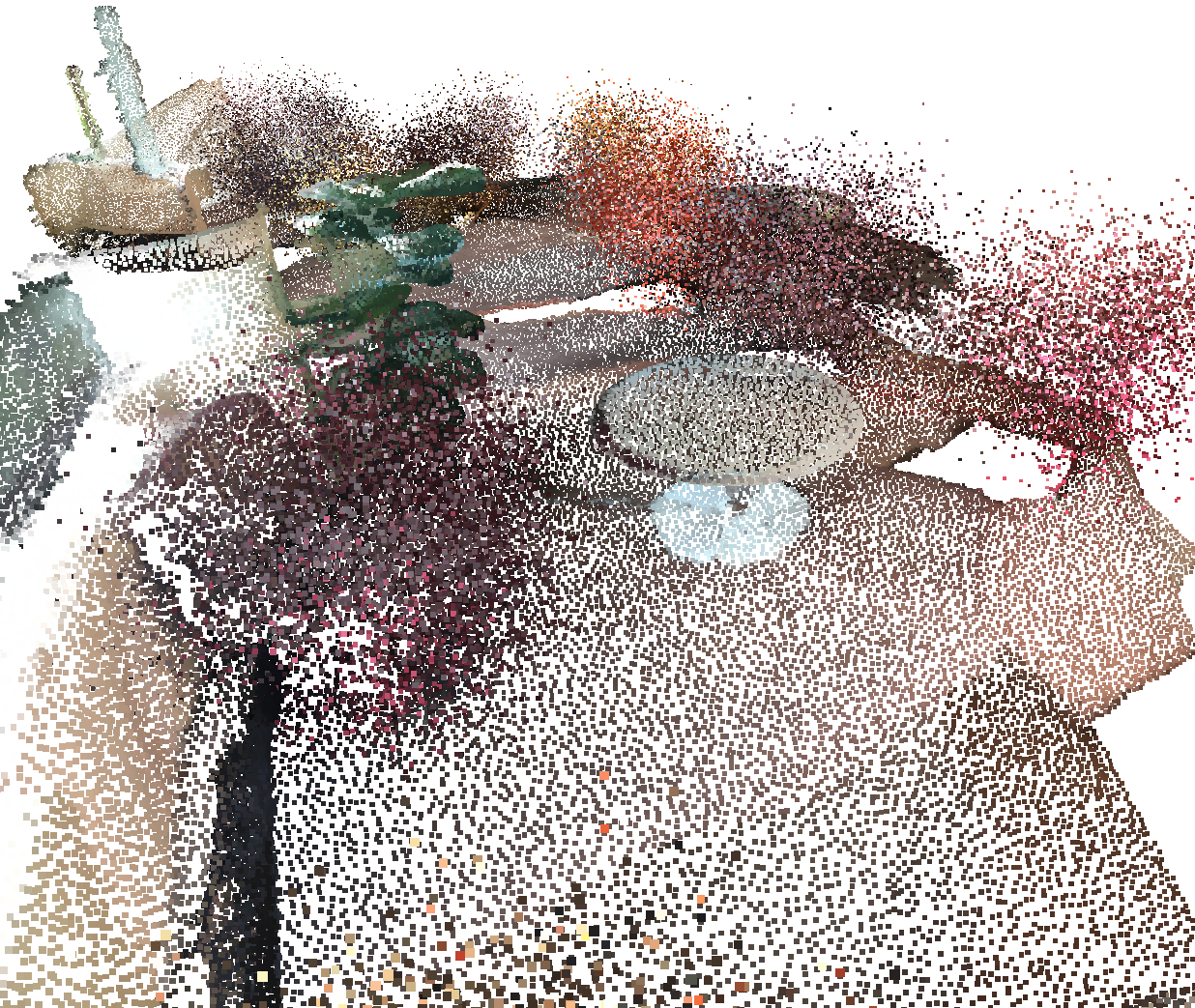}
                 \caption{Local Jitter}
             \end{subfigure}
             \hfill
             \begin{subfigure}[b]{0.24\textwidth}
                 \centering
                 \includegraphics[width=\textwidth]{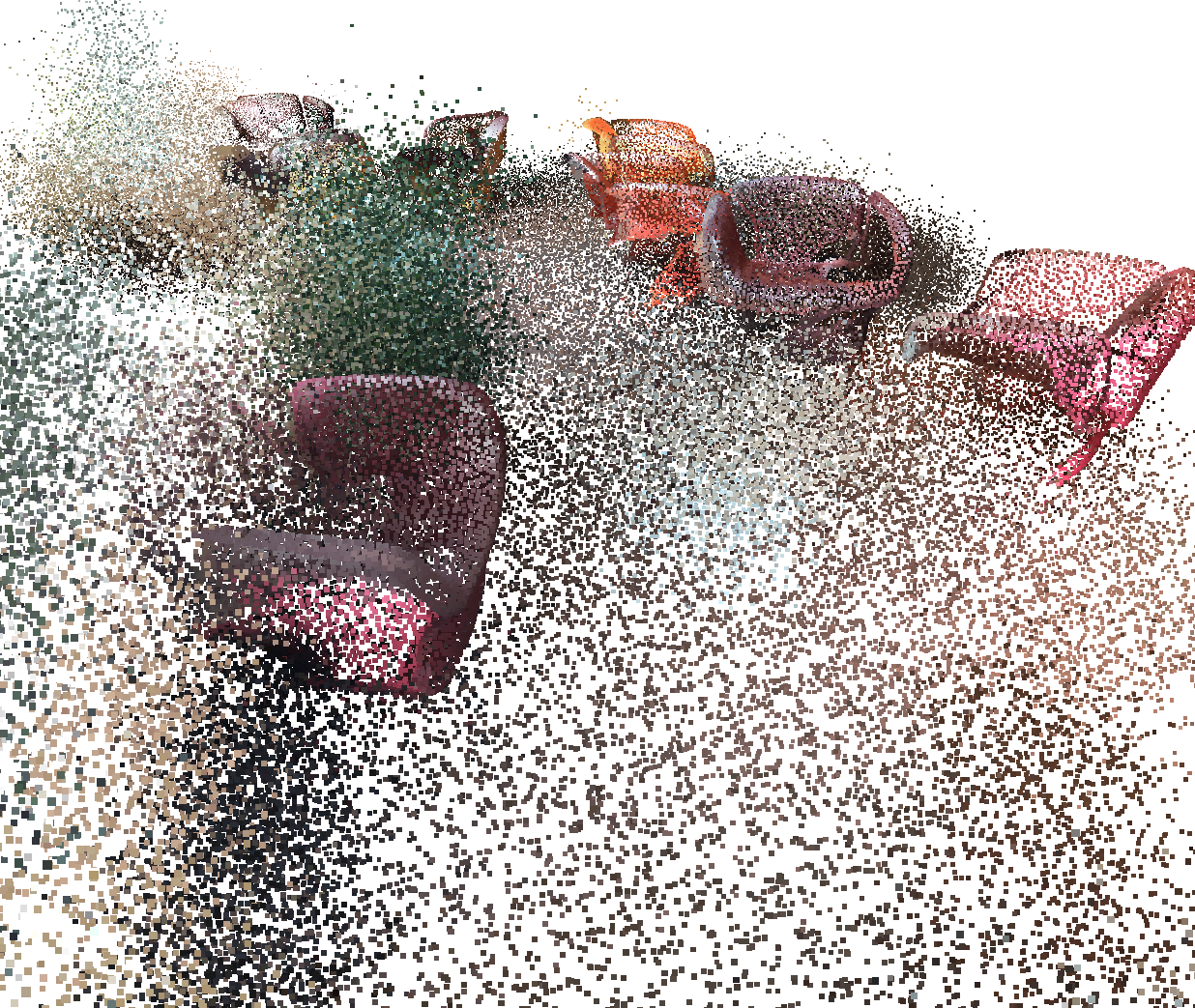}
                 \caption{Bckgrnd Jitter}
             \end{subfigure}\vspace{-0.1cm}
                \caption{An example scene with its corrupted versions (Point Jitter, Local Jitter, and Background Jitter). }\vspace{-0.15cm}
                \label{fig:jitter}
        \end{figure}
        
        \begin{figure}
            \includegraphics[width=\textwidth]{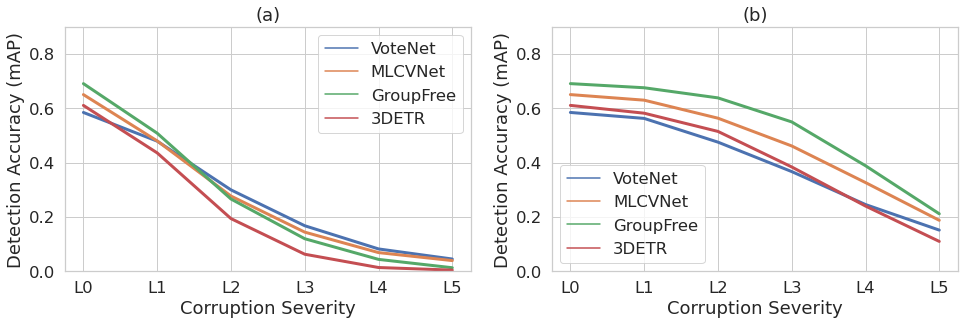}
            \caption{Detection Accuracy (mAP@0.25 IoU) for (a) Jitter and (b) Floor Plane Inclination corruptions over different architectures.} \vspace{-0.4cm}
            \label{jitter-rotation_graph}
        \end{figure}
        
        \subsubsection{Point Jitter.}
        In Jitter corruption, we normalize the scene then add a Gaussian noise $\epsilon \in \mathcal{N}(0, \sigma^2)$ to (X,Y,Z) coordinates for all scene points.  We set $\sigma = 0.004 * level$.
        Fig. \ref{jitter-rotation_graph}a shows the performance of all detectors over 5 levels of Jitter. VoteNet shows good robustness against jitter, most probably due to its local grouping nature.  

        \subsubsection{Floor Plane Inclination.}
        In this corruption, we want to investigate the robustness of detectors toward orientation changes and noise. This scenario might arise in inclined indoor scenes like theatres and stairs. We apply rotation around the Z-axis with different angles ranging from $5$ to $25$ degrees in steps of $5$. 
        As can be seen in Fig. \ref{jitter-rotation_graph}b, while the degree of rotation increases the accuracy worsens across detectors. In the initial levels, Group-Free is superior as it shows more robustness toward inclinations of small angles. 
        
        \begin{table}[h!]
        \vspace{-0.4cm}
        \centering
        \caption{Per-category evaluation on clean validation-set of ScanNet evaluated with AP@0.25 IoU.}\vspace{-0.2cm}
        \label{tab:percat1}
        \resizebox{\textwidth}{!}{%
        \begin{tabular}{|l|cccccccccccccccccc|c|}
        \hline
        Method & cab & bed & chair & sofa & tabl & door & windw & bkshf & pic. &         contr & desk & curtn & frdge & showr & toilt & sink & bath & ofurn & mAP \\
        \hline
        VoteNet\cite{votenet2019} & 35.00 &	88.60 &	88.88 &	88.07 &	58.53 &	47.46&	36.33&	47.65&	5.70&	59.98&	67.18&	45.13&	45.65&	63.28&	92.64&	53.29&	89.21&	39.37&	58.44 \\
        MLCVNet\cite{mlcvnet2020} & 43.47&	90.22&	90.65&	87.11&	67.44&	55.11&	45.39&	60.03&	13.34&	58.22&	75.04&	54.29&	61.27&	72.00&	99.26&	57.76&	91.26&	48.38&	65.01 \\
        Group-Free\cite{groupfree3d2021} & 53.43&	89.35&	93.17&	86.48&	72.73&	60.24&	52.11&	62.43&	15.69&	69.02&	81.93&	66.38&	46.55&	75.47&	94.63&	74.38&	94.87&	54.00&	69.05 \\
        3DETR\cite{3detr2021} & 48.14&	83.43&	86.45&	82.71&	63.41&	50.09&	41.67&	53.12&	12.11&	51.06&	73.01&	56.21&	54.88&	49.25&	99.15&	67.75&	82.17&	44.16&	61.04 \\
        \hline
        \end{tabular}}
        \vspace{-0.7cm}
        \end{table}
        
        \begin{table}[h!]
        \centering
        \caption{Per-category evaluation on ScanNet after adding Local Noise evaluated with AP@0.25 IoU.}\vspace{-0.3cm}
        \label{tab:percat2}
        \resizebox{\textwidth}{!}{%
        \begin{tabular}{|l|cccccccccccccccccc|c|}
        \hline
        Method & cab & bed & chair & sofa & tabl & door & windw & bkshf & pic. &         contr & desk & curtn & frdge & showr & toilt & sink & bath & ofurn & mAP \\
        \hline
        VoteNet\cite{votenet2019} & 34.97&	88.38&	58.29&	90.06&	53.10&	48.29&	35.74&	45.05&	7.12&	50.85&	65.33&	47.23&	47.01&	59.66&	94.77&	51.14&	90.75&	34.34&	55.67\\
        MLCVNet\cite{mlcvnet2020} & 44.18&	89.52&	64.15&	88.36&	62.20&	56.68&	43.53&	57.63&	13.69&	56.86&	72.21&	55.00&	61.84&	74.56&	96.17&	59.18&	94.34&	45.62&	63.10 \\
        Group-Free\cite{groupfree3d2021} & 54.47&	89.58&	53.04&	86.39&	63.84&	60.05&	51.35&	61.17&	14.50&	69.92&	71.83&	64.48&	49.63&	75.53&	95.83&	73.62&	94.19&	54.05&	65.75 \\
        3DETR\cite{3detr2021} & 40.98&	84.11&	48.29&	84.52&	50.02&	47.66&	36.71&	48.63&	12.54&	57.04&	57.03&	49.24&	55.84&	54.98&	99.71&	63.28&	85.28&	38.73&	56.37 \\
        \hline
        \end{tabular}}\vspace{-0.5cm}
        \end{table}
        
        \begin{table}[h!]
        \centering
        \caption{Per-category evaluation on ScanNet after adding Background Noise evaluated with AP@0.25 IoU.} \vspace{-0.3cm}
        \label{tab:percat3}
        \resizebox{\textwidth}{!}{%
        \begin{tabular}{|l|cccccccccccccccccc|c|}
        \hline
        Method & cab & bed & chair & sofa & tabl & door & windw & bkshf & pic. &         contr & desk & curtn & frdge & showr & toilt & sink & bath & ofurn & mAP \\
        \hline
        VoteNet\cite{votenet2019} & 1.35&	21.27&	16.00&	1.42&	10.59&	1.02&	3.96&	2.12&	0.00&	3.77&	9.51&	1.84&	1.58&	0.05&	1.26&	1.08&	4.15&	0.33&	4.52\\
        MLCVNet\cite{mlcvnet2020} & 0.49 &	20.68&	10.42&	1.68&	4.01&	0.27&	1.58&	1.14&	0.00&	1.72&	7.31&	0.70&	2.81&	0.29&	8.20&	1.05&	8.06&	0.54&	3.94 \\
        Group-Free\cite{groupfree3d2021} &0.35&	6.22&	3.06&	2.28&	3.91&	0.44&	1.05&	0.93&	0.00&	1.52&	1.92&	1.15&	0.10&	0.00&	0.21&	0.34&	0.00&	0.06&	1.31 \\
        3DETR\cite{3detr2021} &0.17&	2.01&	0.68&	0.31&	0.35&	2.21&	0.23&	0.72&	0.00&	0.00&	0.57&	0.07&	0.11&	0.01&	0.05&	0.00&	0.34&	0.04&	0.44 \\
        \hline
        \end{tabular}}\vspace{-0.6cm}
        \end{table}
        
        \subsubsection{Local Noise.}
        How large is the effect of context on object detection? In Local Noise, our goal is to answer this question. We distort the shape information of the object by replacing all instances of an object class in all scenes with random noise. We choose chair class as it is the most occurring instance with 1368 chair instances out of 4364 total instances. Here, we apply the previously defined Jitter corruption with a severity level of 5 to the chair points. 
        For comparison, Table \ref{tab:percat1} shows per category AP over the clean validation-set of ScanNet. In Table \ref{tab:percat2} we report per category AP after applying Local Noise corruption. The AP of the class chair drops as expected. In particular, Group-Free chair's AP drops from $93.17$ to $53.04$. We suspect the sharper drop is due to the noise affecting the self-attention modules. In the other hand, MLCVNet performs the best. The added context modules helped the model recognize chair instances from other objects. 
        Furthermore, for all detectors, the AP of the table category decreases significantly. We reason that behind this drop is noisy points interfering with the table points.  
        
        \subsubsection{Background Noise.} 
        Introducing background noise aims to investigate how the detector performance would be affected if there is no useful context information but rather noise around the object. Here, we keep the local object information intact. We apply Jitter (level 5) on all points except class chair points.
        As shown in Table \ref{tab:percat3}, all category's AP deteriorate. Although the local shape information of the chair is preserved, all detectors fail to detect it with certainty. We can see that VoteNet performs the best with AP of 16\%, while 3DETR performs the worst with AP of less than 1 \%. This aligns with our previous observations of VoteNet's superior robustness to noise and 3DETR extreme sensitivity to noise.  
        
        \begin{table}[h!]
        \vspace{-0.6cm}
        \centering
        \caption{Corruption Error(CE) and mean Corruption Error(mCE) across Points Alterations corruptions in all architectures.} \vspace{-0.3cm}
        \label{tab:CE-PointsAlter}
        \begin{tabular}{l|c|c|c|c|c|c}
        \hline
         Architectures & mAP $\uparrow$ & mCE $\downarrow$ & Jitter &         Floor-P-Incl. & Local-N & Background-N \\
        \hline
        VoteNet\cite{votenet2019} & 58.44 & 1.000 & 1.000 & 1.000 & 1.000 & 1.000          \\
        MLCVNet\cite{mlcvnet2020} & 65.01 & 0.897 & 1.016	& 0.8854 &	0.951 &	1.010 \\
        Group-Free\cite{groupfree3d2021} & 69.05 & 0.842 & 1.031 & 0.7933	& 0.951 & 1.028  \\
        3DETR\cite{3detr2021} & 61.04 & 1.067 & 1.092	& 0.9911 & 1.024 & 1.049 \\
        \hline
        \end{tabular} \vspace{-0.3cm}
        \end{table}
    
    
    CE and mCE of Points Alteration are shown in Table \ref{tab:CE-PointsAlter}. All models perform unfavorably in Jitter and Background Noise. MLCVNet struggles with Jitter and Background Noise but performs relatively better in Local Noise and Floor Plane Inclination. Group-Free follows a similar trend. Furthermore, 3DETR scores the highest CE in Local Noise which is consistent with our previous findings.

\section{Conclusions}

We investigate the robustness of four different 3D detection architectures to 13 point cloud corruptions that cover point addition, point removal, and 3D scene alteration. We identify common natural errors and imperfections in point clouds and then design corruptions that simulate those inaccuracies. We present our experiments and make comparisons across corruptions and different architectures. we hope the analysis presented in this work will contribute toward designing more robust 3D object detection frameworks in the future.

%
%
%
%

\bibliographystyle{splncs04}
\bibliography{egbib}

\end{document}